\documentclass[conference]{IEEEtran}
\IEEEoverridecommandlockouts
\usepackage[utf8]{inputenc}
\usepackage[T1]{fontenc}
\usepackage{blindtext}
\usepackage{amsmath}
\usepackage{amssymb}
\usepackage{amsfonts}
\usepackage{multirow}
\usepackage{caption}
\usepackage{subcaption}
\usepackage{graphicx}
\usepackage{booktabs}
\usepackage{placeins}
\usepackage{tabularx}
\usepackage{url}

\newcolumntype{C}{>{\small\centering\arraybackslash}X}
\newcolumntype{T}{>{\hsize=1.\hsize}X}
\def\BibTeX{{\rm B\kern-.05em{\sc i\kern-.025em b}\kern-.08em
    T\kern-.1667em\lower.7ex\hbox{E}\kern-.125emX}}
\begin{document}

\title{PPG-to-ECG Signal Translation for \\ Continuous Atrial Fibrillation Detection via \\ Attention-based Deep State-Space Modeling}

\author{Khuong Vo$^{1,*}$, Mostafa El-Khamy$^{2}$, and Yoojin Choi$^{2}$
\thanks{$^{1}$Department of Computer Science, University of California, Irvine}
\thanks{$^{2}$Samsung Device Solutions Research America}
\thanks{*Correspondence email: khuongav@uci.edu}
}

\maketitle

\begin{abstract}
Photoplethysmography (PPG) is a cost-effective and non-invasive technique that utilizes optical methods to measure cardiac physiology. PPG has become increasingly popular in health monitoring and is used in various commercial and clinical wearable devices. Compared to electrocardiography (ECG), PPG does not provide substantial clinical diagnostic value, despite the strong correlation between the two. Here, we propose a subject-independent attention-based deep state-space model (ADSSM) to translate PPG signals to corresponding ECG waveforms. The model is not only robust to noise but also data-efficient by incorporating probabilistic prior knowledge. To evaluate our approach, 55 subjects' data from the MIMIC-III database were used in their original form, and then modified with noise, mimicking real-world scenarios. Our approach was proven effective as evidenced by the PR-AUC of 0.986 achieved when inputting the translated ECG signals into an existing atrial fibrillation (AFib) detector. ADSSM enables the integration of ECG's extensive knowledge base and PPG's continuous measurement for early diagnosis of cardiovascular disease.
\end{abstract}


\section{Introduction}
The measurement of the electrical activity generated by an individual's heart, known as an electrocardiogram (ECG), typically requires the placement of several electrodes on the body. ECG is considered the preferred method for monitoring vital signs and for the diagnosis, management, and prevention of cardiovascular diseases (CVDs), which are a leading cause of death globally, accounting for approximately 32\% of all deaths in 2017 according to Global Burden of Disease reports \cite{ref:Allen}. It has also been demonstrated that sudden cardiac arrests are becoming more prevalent in young individuals, including athletes ~\cite{ref:sudden}. Regular ECG monitoring has been found to be beneficial for the early identification of CVDs ~\cite{rosiek2016risk}. Among heart diseases, atrial fibrillation (AFib) is adults' most common rhythm disorder. Identifying AFib at an early stage is crucial for the primary and secondary prevention of cardioembolic stroke, as it is the leading risk factor for this type of stroke \cite{olier2021machine}.
Advancements in electronics, wearable technologies, and machine learning have made it possible to record ECGs more easily and accurately, and to analyze large amounts of data more efficiently. Despite these developments, there are still challenges associated with continuously collecting high-quality ECG data over an extended period, particularly in everyday life situations. 
The 12-lead ECG, considered the clinical gold standard, and simpler versions, such as the Holter ECG, can be inconvenient and bulky due to the need to place multiple electrodes on the body, which can cause discomfort. Additionally, the signals may degrade over time as the impedance between the skin and electrodes changes. Consumer-grade products such as smartwatches have developed solutions to address these issues. However, these products require users to place their fingers on the watch to form a closed circuit, making continuous monitoring impossible.

\begin{figure*}[h!]
\centering
\includegraphics[scale=.35]{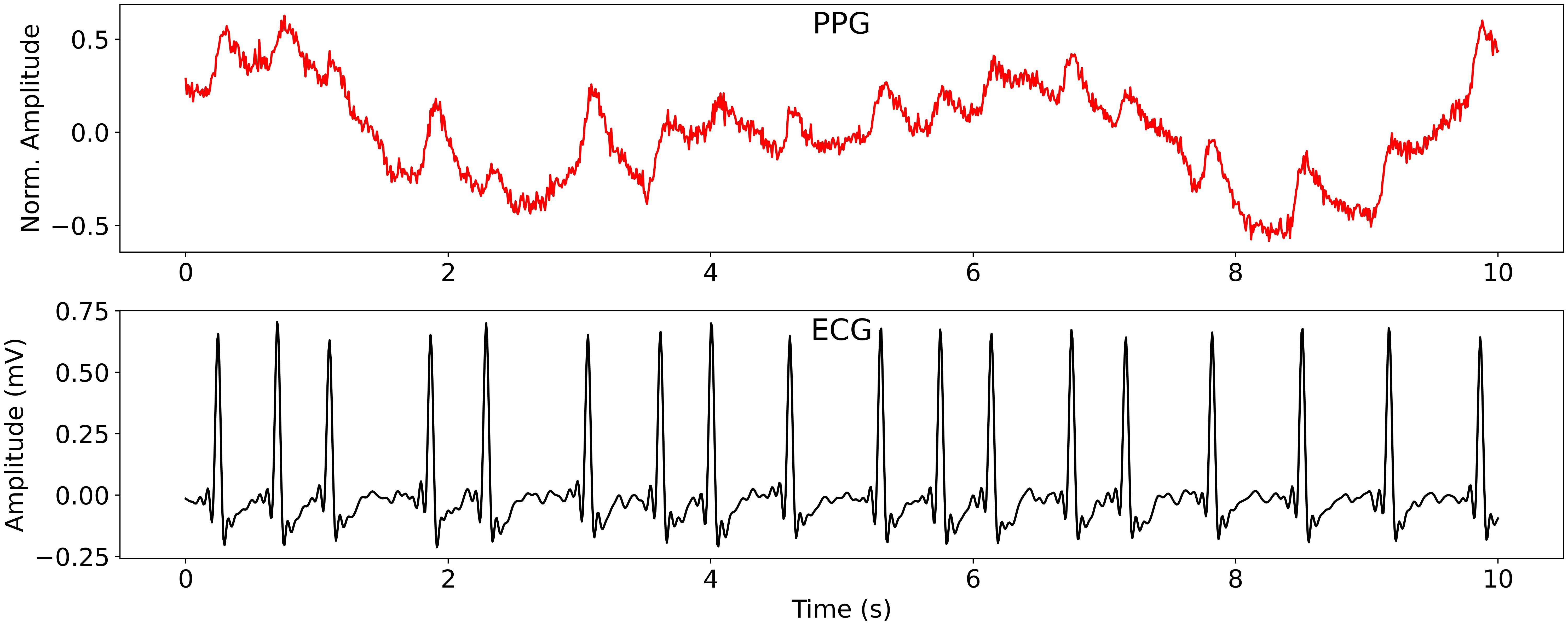}
\caption{A PPG-ECG waveform pair. PPG signals can often become contaminated by noise.}
\label{fig:preprocessed}
\end{figure*}

One potential solution to these issues is to use a mathematical method to derive ECG data from an alternative, highly correlated, non-invasive signal, such as the photoplethysmogram (PPG), which can be easily acquired using various wearable devices, including smartwatches. 
PPG is more convenient, cost-effective, and user-friendly. PPG has been increasingly adopted in consumer-grade devices. This technique involves the use of a light source, usually an LED, and a photodetector to measure the changes in light absorption or reflection as blood flows through the tissue. 
ECG and PPG signals are inherently correlated as both are influenced by the same underlying cardiac activity, namely the depolarization and repolarization of the heart. These contractions lead to changes in peripheral blood volume, which are measured by PPG.  
Figure \ref{fig:preprocessed} shows the relationship between ECG and PPG waveforms.
Although there are established standards for interpreting ECG for clinical diagnosis, the use of PPG is still mostly limited to measuring heart rate and oxygen saturation ~\cite{reisner2008utility}. By translating PPG to ECG signals, clinical diagnoses of cardiac diseases and anomalies could be made in real-time.

Few research works attempted to synthesize ECG from PPG signals. In \cite{banerjee2014photoecg}, a machine learning-based approach was proposed to estimate the ECG parameters, including the RR, PR, QRS, and QT intervals, using features from the time and frequency domain extracted from a fingertip PPG signal. Additionally, \cite{zhu2019ecg, tian2020cross} proposed models to reconstruct the entire ECG signal from PPG in the frequency domain. However, the performance of these approaches relied on cumbersome algorithms for feature crafting. With recent advances in deep learning, \cite{vo2021p2e, sarkar2021cardiogan, chiu2020reconstructing} leveraged the expressiveness and structural flexibility of neural networks to build end-to-end PPG-to-ECG algorithms. 
However, the models suffer from data-hungry problems as they do not explicitly model the underlying sequential structures of the data. 
In addition, complex deep learning models cannot run efficiently on resource-constrained devices (e.g., wearables) due to their high computational intensity, which poses a critical challenge for real-world deployment \cite{lee2020stint}.
Furthermore, deterministic models face difficulties in effectively generalizing to noisy data.

To address these challenges, we propose a deep probabilistic model to accurately estimate ECG waveforms from raw PPG. The contributions of this work are three-fold:
\begin{itemize}
    \item We present a deep generative model incorporating prior knowledge about the data structures that enable learning on small datasets. Specifically, we develop a deep latent state-space model augmented by an attention mechanism.
    \item The probabilistic nature of the model enhances its robustness to noise. We demonstrate this by evaluating the model on data corrupted with Gaussian and baseline wandering noise, replicating real-life situations.
    \item Our method is effective not only in healthy subjects but also in subjects with AFib. It is orthogonal and complementary to existing AFib detection methods \cite{hong2020opportunities} by simply providing the translated ECG to any pre-trained models. This would enhance the performance of existing models by enabling uninterrupted monitoring, thereby facilitating the early detection of cardiovascular disease.
\end{itemize}

% The rest of our paper is organized as follows. 
% Section 2 presents the dynamical model to translate PPG to ECG signals and the latent factors of data variations.
% In Section 3, experimental data are presented, and results on signal translation and Afib detection performance are discussed. 
% Finally, Section 4 wraps up the paper.

\section{Methodology}

\subsection{State-Space Modeling of ECG from PPG Signals}
% In the previous section, we consider the networks that process the entire time series as a whole, which do not explicitly model the underlying sequential natures of the data. This may lead to resource-inefficient learning.  Here, propose to address the problems by leveraging the  
% \textit{quasi-periodic nature} of the physiological signals.

\subsubsection{ECG Generative (Decoding) Process from PPG}
We consider nonlinear dynamical systems with observations $\mathbf{y}_{t} \in \mathbb{R}^{n_{rr}}$, i.e., RR intervals or the time elapsed between two successive R peaks on the ECG, depending on control inputs $\mathbf{x}_{t} \in \mathbb{R}^{n_{pp}}$, i.e., PP intervals or the time elapsed between two successive systolic peaks on the PPG. We choose the peaks to segment the signals as they are the most robust features. Corresponding discrete-time sequences of length $T$ are denoted as $\mathbf{y}_{1: T}=\left(\mathbf{y}_{1}, \mathbf{y}_{2}, \ldots, \mathbf{y}_{T}\right)$ and $\mathbf{x}_{1: T}=\left(\mathbf{x}_{1}, \mathbf{x}_{2}, \ldots, \mathbf{x}_{T}\right)$.

Given an input PPG $\mathbf{x}_{1: T}$, we are interested in a probabilistic model $p\left(\mathbf{y}_{1: T} \mid \mathbf{x}_{1: T}\right)$. Formally, we consider
\begin{equation}
p\left(\mathbf{y}_{1: T} \mid \mathbf{x}_{1: T}\right)=\int p\left(\mathbf{y}_{1: T} \mid \mathbf{z}_{1: T}, \mathbf{x}_{1: T}\right) p\left(\mathbf{z}_{1: T} \mid \mathbf{x}_{1: T}\right) \mathrm{d} \mathbf{z}_{1: T}
\end{equation}
where $\mathbf{z}_{1: T}$ represents the latent sequence associated with the given model. This implies that we are considering a generative model that incorporates a latent dynamical system with an emission model $p\left(\mathbf{y}_{1: T} \mid \mathbf{z}_{1: T}, \mathbf{x}_{1: T}\right)$ and transition model $p\left(\mathbf{z}_{1: T} \mid \mathbf{x}_{1: T}\right)$.

\begin{figure}[h!]
\centering
\includegraphics[scale=.5]{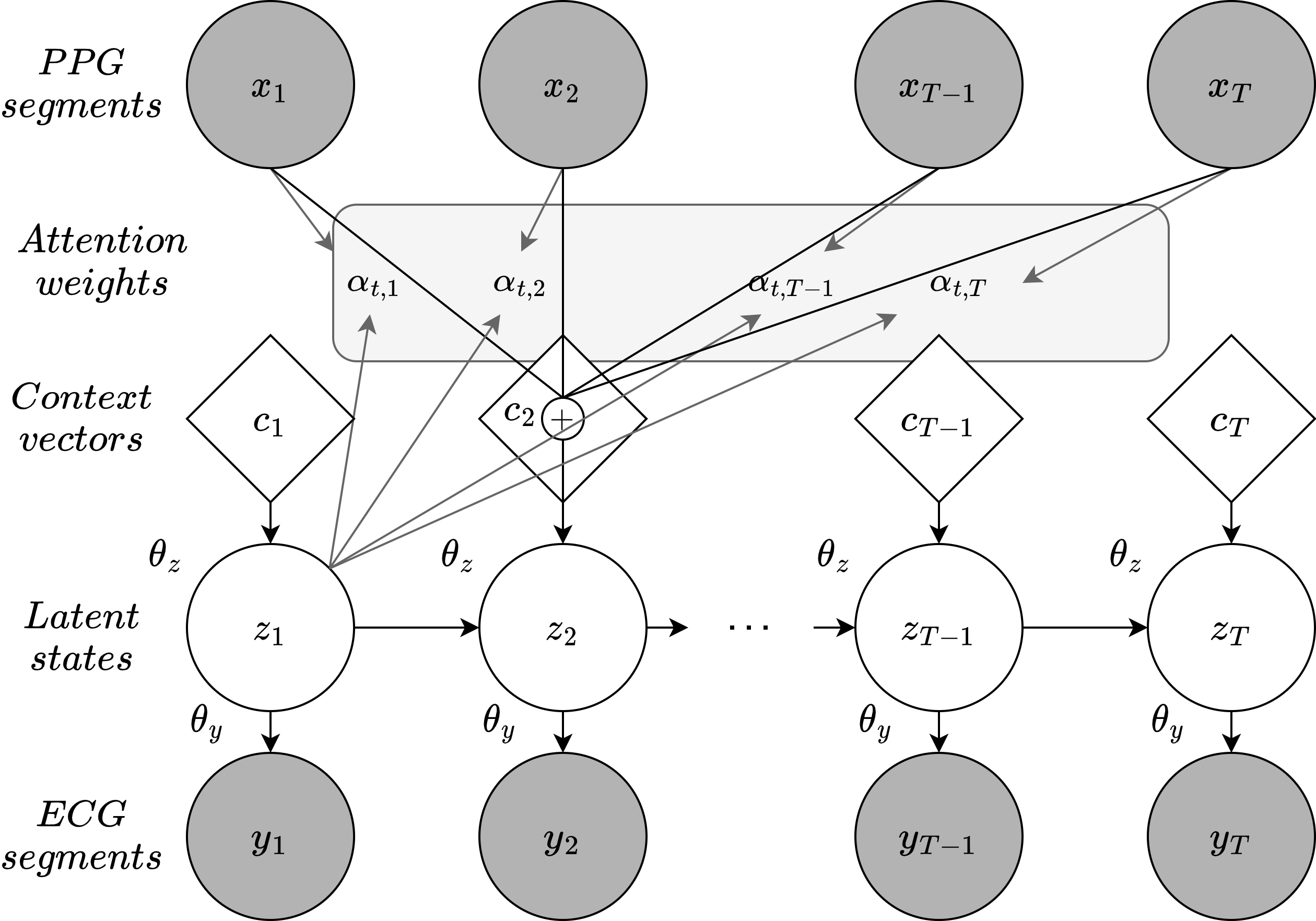}
\caption{The graphical model for ECG translation from PPG. Shaded nodes represent observed variables. Clear nodes represent latent variables. Diamond nodes denote deterministic variables. Variables $\mathbf{x}_t, \mathbf{y}_t$, and $\mathbf{c}_t$ represent PP intervals, RR intervals, and context vectors, respectively. $\alpha_{t,i}$ are attention weights defines how well two intervals $\mathbf{x}_i$ and $\mathbf{y}_t$ are aligned. The attention mechanism is shown only at time step 2.}
\label{fig:gen}
\end{figure}

To derive state-space models, we make certain assumptions regarding the state transition and emission models, as shown in Figure \ref{fig:gen}:
\begin{align}
p\left(\mathbf{z}_{1: T} \mid \mathbf{x}_{1: T}\right) & =\prod_{t=0}^{T-1} p\left(\mathbf{z}_{t+1} \mid \mathbf{z}_{t}, \mathbf{x}_{1: T}\right) \\
p\left(\mathbf{y}_{1: T} \mid \mathbf{z}_{1: T}, \mathbf{x}_{1: T}\right) & =\prod_{t=1}^{T} p\left(\mathbf{y}_{t} \mid \mathbf{z}_{t}\right)
\end{align}
Equations 2 and 3 make the assumption that the current state $\mathbf{z}_{t}$ includes all the relevant information about both the current observation $\mathbf{y}_{t}$ and the next state $\mathbf{z}_{t+1}$, given the current control input $\mathbf{x}_{t}$.

% \newpage

In contrast to the DKF model of \cite{krishnan2015deep, krishnan2017structured}, our model takes into account the entire input signal $\mathbf{x}_{1: T}$ for each output $\mathbf{y}_{t}$ via an attention mechanism \cite{bahdanau2014neural}.
Note that there are usually misalignments between the PPG and ECG cycles. Therefore, it is difficult to construct optimal and exact sample pairs. This attention mechanism not only helps to add more context to generate ECG segments, but also helps to address the problem of misalignment. 
 
Let us define $\mathbf{c}_t$ a sum of features of the input sequence (PP intervals), weighted by the alignment scores:
\begin{align}
\mathbf{c}_t & =\sum_{i=1}^T \alpha_{t, i} \boldsymbol{x}_i \\
\alpha_{t, i} 
% & =\operatorname{align}\left(\boldsymbol{y}_t, \boldsymbol{x}_i\right) \\
& =\frac{\exp \left(\mathbf{s}\left(\mathbf{z}_{t-1}, \boldsymbol{x}_i\right)\right)}{\sum_{i^{\prime}=1}^n \exp \left(\mathbf{s}\left(\mathbf{z}_{t-1}, \boldsymbol{x}_{i^{\prime}}\right)\right)}
\end{align}
The alignment function $\mathbf{s}$ assigns a score $\alpha_{t,i}$ to the pair of input at position $i$ and output at position $t$, ($\mathbf{x}_i, \mathbf{y}_t$), based on how well they match. The set of $\alpha_{t,i}$ are weights defining how much of each source segment should be considered for each output interval. 

Both state transition (prior) and emission models are non-linear Gaussian transformations parametrized by neural networks $\theta_z$ and $\theta_y$ :
\begin{align}
 p_{\theta_z}(\mathbf{z}_{t+1} \mid \mathbf{z}_t, \mathbf{x}_{1: T}) & = \mathcal{N}(\mathbf{z}_{t+1} \mid \boldsymbol{\mu}_{\theta_z}(\mathbf{z}_t, \mathbf{c}_{t+1}), \boldsymbol{\sigma}^2_{\theta_z}(\mathbf{z}_t, \mathbf{c}_{t+1})); \\
 p_{\theta_y}(\mathbf{y}_t \mid \mathbf{z}_t) & =\mathcal{N}(\mathbf{y}_t \mid \boldsymbol{\mu}_{\theta_y}(\mathbf{z}_t), \mathbf{I})
\end{align}

where $\boldsymbol{\mu}$ and $\boldsymbol{\sigma}^2$ are the means and diagonal covariance matrices of the normal distributions $\mathcal{N}$, $\mathbf{I}$ is the identity covariance matrix.

\subsubsection{Latent State Inference (Posterior Encoding) Process}

\begin{figure}[h!]
\centering
\includegraphics[scale=.5]{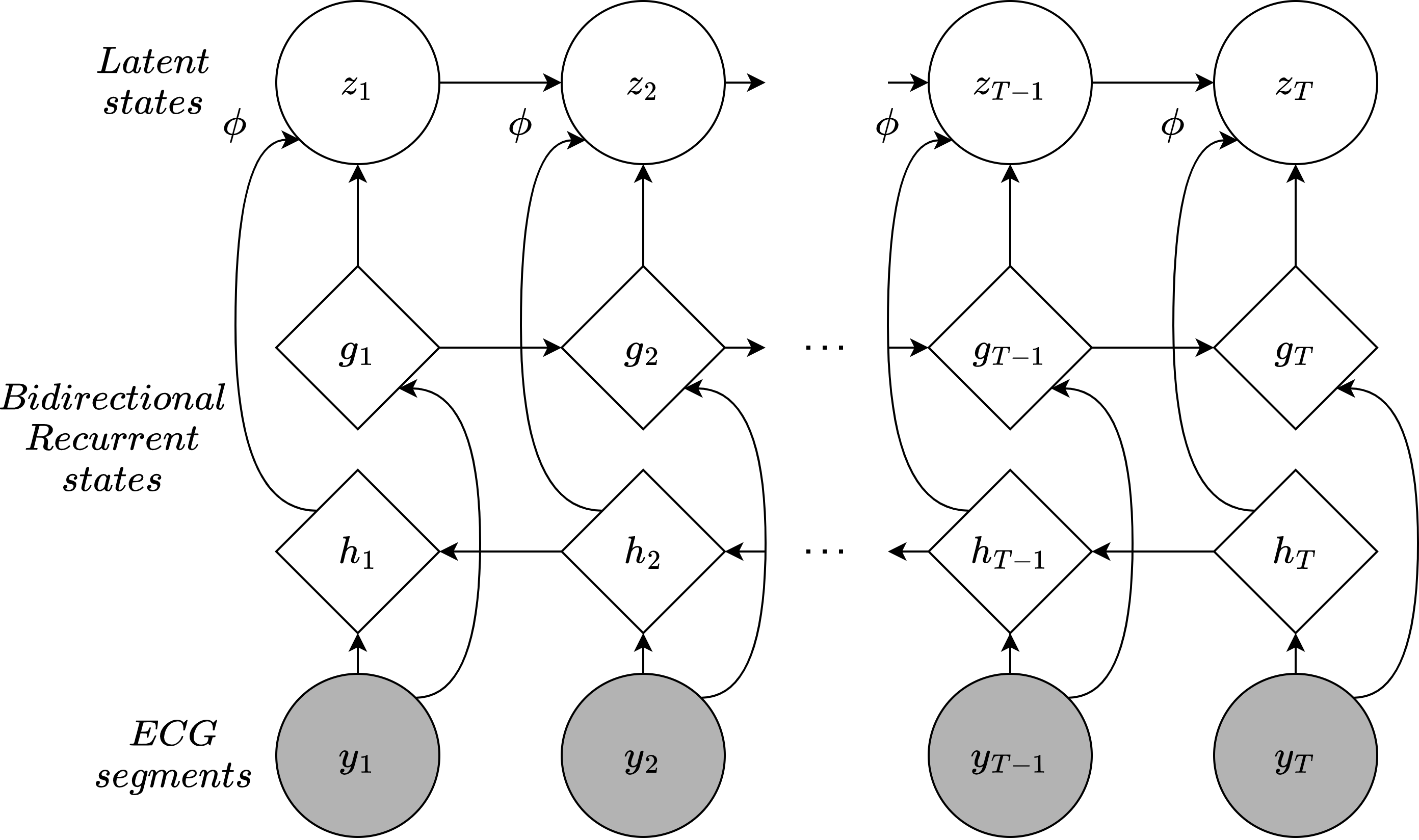}
\caption{The graphical model at latent state inference time. Variables $\mathbf{y}_t, \mathbf{h}_t, \mathbf{g}_t$, and $\mathbf{z}_t$ represent respectively RR intervals, backward, forward recurrent states, and latent states.}
\label{fig:infer}
\end{figure}

Unlike a deterministic translation model, the process needs to find meaningful probabilistic embeddings of ECG segments in the latent space. We want to identify the structure of the parametetrized posterior distribution $q_\phi\left(\mathbf{z}_{1: T} \mid \mathbf{y}_{1: T}\right)$. Notice that we made a design choice to perform inference using only $\mathbf{y}_{1: T}$. We chose this with the conditional independence assumption that PPG segments do not provide more information than ECG segments alone. The graphical model in Figure \ref{fig:gen} shows that the $\mathbf{z}_{t}$ node blocks all information coming from the past and flowing to $\mathbf{z}_{t+1}$ (i.e., $\mathbf{z}_{1: t-1}$ and $\left.\mathbf{y}_{1: t}\right)$, leading to the following structure as in Figure \ref{fig:infer}:
\begin{equation}
q_\phi\left(\mathbf{z}_{1: T} \mid \mathbf{y}_{1: T}\right)=q_\phi\left(\mathbf{z}_{1} \mid \mathbf{y}_{1: T}\right)\prod_{t=1}^{T-1} q_\phi\left(\mathbf{z}_{t+1} \mid \mathbf{z}_{t}, \mathbf{y}_{t+1: T}\right)    
\end{equation}
where
\begin{equation}
 q_{\phi}(\mathbf{z}_{t+1} \mid \mathbf{z}_t, \mathbf{y}_{t+1: T}) = \mathcal{N}(\mathbf{z}_{t+1} \mid \boldsymbol{\mu}_{\phi}(\mathbf{z}_t, \mathbf{y}_{t+1: T}), \boldsymbol{\sigma}^2_{\phi}(\mathbf{z}_t, \mathbf{y}_{t+1: T}))
\end{equation}

\subsubsection{Training Process}
The objective function becomes a timestep-wise conditional variational lower bound \cite{kingma2013auto, sohn2015learning, krishnan2017structured}: 
\begin{equation}
\begin{aligned}
& \log p_\theta(\mathbf{y} \mid \mathbf{x}) \geq \mathcal{L}(\mathbf{x}, \mathbf{y} ;\theta_y, \theta_z, \phi) \triangleq \\
& \sum_{t=1}^T \underset{q_\phi\left(\mathbf{z}_t \mid \mathbf{y}_{t:T}\right)}{\mathbb{E}}[\overbrace{\log \underbrace{p_{\theta_y}\left(\mathbf{y}_t \mid \mathbf{z}_t\right)}_{\text {emission model}}}^{\text {reconstruction}}] \\ 
& - \beta\overbrace{\operatorname{KL}\left(q_\phi\left(\mathbf{z}_1 \mid \mathbf{y}_{1:T}\right) \| p_{\theta_z}\left(\mathbf{z}_1 \mid \mathbf{x}_{1:T}\right)\right)}^{\text {regularization}} \\
& -\beta\sum_{t=1}^{T-1} \underset{q_\phi\left(\mathbf{z}_{t} \mid \mathbf{y}_{t:T}\right)}{\mathbb{E}}[\overbrace{\operatorname{KL}(\underbrace{q_\phi\left(\mathbf{z}_{t+1} \mid \mathbf{z}_{t}, \mathbf{y}_{t:T}\right)}_{\text {posterior inference model}}|| \underbrace{p_{\theta_z}\left(\mathbf{z}_{t+1} \mid \mathbf{z}_t, \mathbf{x}_{1:T}\right)}_{\text {prior transition model}})}^{\text {regularization}}]
\end{aligned}
\end{equation}
where $\beta$ controls the regularization strength. During training, the Kullback–Leibler (KL) losses in the regularization terms "pull" the posterior distributions (which encode EEG segments) and the prior distributions (which embed PPG segments) towards each other. We learn the generative and inference models jointly by maximizing the conditional variational lower bound with respect to their parameters.
\subsection{Neural Network Architectures}
Let us denote $\mathbf{W}$, $\mathbf{v}$, and $\mathbf{b}$ the weight matrices.
% \indent\indent 

\textbf{Score Model}: The alignment score $\alpha$ in Equation 5 is parametrized by a feedforward network with a single hidden layer, and this network is jointly trained with other parts of the model. 
The score function $\mathbf{s}$ is in the following form:
\begin{equation}
    \mathbf{s}\left(\boldsymbol{z}_{t-1}, \boldsymbol{x}_i\right)=\mathbf{v}_s^{\top} \tanh \left(\mathbf{W}_{s}\left[\mathbf{z}_{t-1}; \mathbf{W}_{x}\boldsymbol{x}_i\right] + \mathbf{b}_s \right)
\end{equation}
% where both $\mathbf{v}_s$ and $\mathbf{W}_s$ are weight matrices to be learned in the alignment model.
\textbf{Prior Transition Model}: We parametrize the transition function in Equation 6 from $z_t$ to $z_{t+1}$ using a Gated Transition Function as in \cite{krishnan2017structured}. The model is flexible in choosing a non-linear transition for some dimensions while having linear transitions for others. The function is parametrized as follows:
\begin{equation}
\begin{aligned}
% & \mathbf{g}_t=\operatorname{MLP}\left(\left[\mathbf{z}_{t}; \mathbf{c}_{t+1}\right], \operatorname{ReLU},\right. \text { sigmoid) (Gating Unit) } \\
& \mathbf{g}_t= \operatorname{sigmoid}(\\ & \mathbf{W}_{g_3}\operatorname{ReLU} \left(\mathbf{W}_{g_2}\operatorname{ReLU} \left(\mathbf{W}_{g_1}\left[\mathbf{z}_{t}; \mathbf{c}_{t+1}\right] + \mathbf{b}_{g_1}\right) + \mathbf{b}_{g_2}\right) + \mathbf{b}_{g_3})\\
% & \mathbf{h}_t=\operatorname{MLP}\left(\left[\mathbf{z}_{t}; \mathbf{c}_{t+1}\right], \operatorname{ReLU}, \mathbb{I}\right) \text { (Proposed mean) } \\
&  \mathbf{d}_t=  \\ & \mathbf{W}_{d_3}\operatorname{ReLU}\left(\mathbf{W}_{d_2}\operatorname{ReLU} \left(\mathbf{W}_{d_1}\left[\mathbf{z}_{t}; \mathbf{c}_{t+1}\right] + \mathbf{b}_{d_1}\right) + \mathbf{b}_{d_2}\right) + \mathbf{b}_{d_3}\\
& \boldsymbol{\mu}_{\theta_z}(\mathbf{z}_t, \mathbf{c}_{t+1})=\left(1-\mathbf{g}_t\right) \odot\left(\mathbf{W}_{\mu_z} \left[\mathbf{z}_{t}; \mathbf{c}_{t+1}\right]+\mathbf{b}_{\mu_z}\right)+\mathbf{g}_t \odot \mathbf{d}_t \\
& \boldsymbol{\sigma}^2_{\theta_z}(\mathbf{z}_t, \mathbf{c}_{t+1})=\operatorname{softplus}\left(\mathbf{W}_{\sigma_z^2} \operatorname{ReLU}\left(\mathbf{d}_t\right)+\mathbf{b}_{\sigma_z^2}\right)
\end{aligned}
\end{equation}
where $\mathbb{I}$ denotes the identity function, and $\odot$ denotes element-wise multiplication.

\textbf{Emission Model}: We parameterize the emission function in Equation 7 using a two-hidden layer network as:
\begin{equation}
\begin{aligned} 
& \boldsymbol{\mu}_{\theta_y}\left(\mathbf{z}_t\right)= \\ & \mathbf{W}_{e_3} \operatorname{ReLU} \left(\mathbf{W}_{e_2}\operatorname{ReLU} \left(\mathbf{W}_{e_1}\mathbf{z}_t + \mathbf{b}_{e_1}\right) + \mathbf{b}_{e_2}\right) + \mathbf{b}_{e_3}
\end{aligned}
\end{equation}

\textbf{Posterior Inference Model}: We use a Bi-directional Gated Recurrent Unit network \cite{chung2014empirical} (GRU) to process the sequential order of RR intervals backward from $\mathbf{y}_T$ to $\mathbf{y}_{t+1}$ and forward from $\mathbf{y}_{t+1}$ to $\mathbf{y}_{T}$. The GRUs are denoted here as $\boldsymbol{h}_{t}=\operatorname{GRU}\left(\mathbf{W}_y\boldsymbol{y}_{T}, \ldots, \mathbf{W}_y\boldsymbol{y}_{t+1}\right)$ and $\boldsymbol{g}_{t}=\operatorname{GRU}\left(\mathbf{W}_y\boldsymbol{y}_{t+1}, \ldots, \mathbf{W}_y\boldsymbol{y}_{T}\right)$, respectively. The hidden states of the GRUs parametrize the variational distribution, which are combined with the previous latent states for the inference in Equation 9 as follows:
\begin{equation}
\begin{aligned}
\mathbf{\tilde{h}}_{t} & =\frac{1}{3}\left(\tanh \left(\mathbf{W}_{h} \mathbf{z}_{t}+\mathbf{b}_{h}\right)+\mathbf{h}_t +  \mathbf{g}_t\right) \\
\boldsymbol{\mu}_{\phi}(\mathbf{z}_t, \mathbf{y}_{t+1: T}) & =\mathbf{W}_\mu \mathbf{\tilde{h}}_{t}+\mathbf{b}_\mu \\
\boldsymbol{\sigma}^2_{\phi}(\mathbf{z}_t, \mathbf{y}_{t+1: T}) & =\operatorname{softplus}\left(\mathbf{W}_{\sigma^2} \mathbf{\tilde{h}}_{t}+\mathbf{b}_{\sigma^2}\right) 
\end{aligned}
\end{equation}

All the hidden layer sizes are 256, and the latent space sizes are 128. Input and output segments at each timestep are of size 90. We use Adam \cite{kingma2014adam} for optimization, with a learning rate of 0.0008, exponential decay rates $\beta_1$ = 0.9, and $\beta_2$ = 0.999. We train the models for 5000 epochs, with a minibatch size 128. We set the regularization hyperparameter $\beta = 0$ at the beginning of training and gradually increase it until $\beta = 1$ is reached at epoch 1250.

\section{Experiments}

\subsection{Dataset}
The MIMIC-III Waveform Database Matched Subset \cite{moody2020mimic, johnson2016mimic} was used for the experiments. The database contains recordings collected from patients at various hospitals. Each session has multiple physiological signals, including PPG and ECG signals, sampled at a frequency of 125 Hz. We used the records of 43 healthy subjects and 12 subjects having AFib, including 30 males and 25 females, 23-84 years old. The dataset is made publicly available \footnote{https://github.com/khuongav/dvae\_ppg\_ecg}. Each record duration is 5 minutes. The first 48 s of each record were used as the training set, the next 12 s as the validation set, and the remaining 228 s as the test set. The preprocessing steps, including filtering, alignment, and normalization, were performed as described in \cite{tang2022robust}. We applied HeartPy \cite{van2019heartpy, van2019analysing} to identify peaks in PPG signals. Each long signal is split into 4-s chunks.
Each peak-to-peak interval was linearly interpolated to a length of 90 during training, which is the mean length of the intervals in the training set. The original interval length information can be preserved by making it an additional feature along with each normalized interval. Alternatively, we can apply padding instead of interpolation. However, we found that these did not contribute to improving the performance under the experimental setting. Original PP interval lengths were used as RR interval lengths in translated ECG signals during testing. This can be justified, as PPG recordings are used to analyze heart rate variability as an alternative to ECG \cite{lu2009comparison, aschbacher2020atrial}. Noise was added to the signals for robustness evaluation. The amplitudes of the baseline noise signals are 0.3, 0.4, and 0.1, and the frequencies are 0.3, 0.2 and 0.9 Hz, respectively. Gaussian noise of standard deviation 0.3.
% Figure \ref{fig:preprocessed} shows an example of the preprocessed with noise-added PPG-ECG waveform pairs.

\subsection{Evaluation Metrics}
\subsubsection{ECG Translation from PPG}
\indent\indent {Pearson's correlation coefficient} \(\left(\rho\right)\) measures how much an original ECG signal \( \mathbf{y}_{1:T} \) and its reconstruction \( \hat{\mathbf{y}}_{1:T} \) co-vary:
% \(\rho\) ranges from -1 to 1, where the magnitude of \(\rho\) indicates the strength of the correlation and the sign of \(\rho\) determines whether the correlation is positive or negative. 
\begin{equation}
\rho=\frac{\left(\mathbf{y}_{1:T}-\bar{y}_{1:T}\right)^{\top}(\hat{\mathbf{y}}_{1:T}-\bar{\hat{y}}_{1:T})}{\left\|\mathbf{y}_{1:T}-\bar{y}_{1:T}\right\|_2\|\hat{\mathbf{y}}_{1:T}-\bar{\hat{y}}_{1:T}\|_2}
\end{equation}
% where \( \mathbf{y}_{1:T} \) and \( \hat{\mathbf{y}}_{1:T} \) represent the original ECG and its reconstruction respectively.

{Root Mean Squared Error} (RMSE) measures the differences between the values of the original signal and its reconstruction:
% The closer the RMSE is to zero, the more accurate the reconstruction.
\begin{equation}
    \text{RMSE}=\frac{\left\|\mathbf{y}_{1:T}-\hat{\mathbf{y}}_{1:T}\right\|_2}{\sqrt{n_y}}
\end{equation}

{Signal-to-Noise Ratio} (SNR) compares the level of the desired signal to the level of undesired noise:

\begin{equation}
    \text{SNR}=20\log\frac{\left\|\mathbf{y}_{1:T}\right\|_2^2}{\left\|\mathbf{y}_{1:T} - \hat{\mathbf{y}}_{1:T}\right\|_2^2}
\end{equation}

\subsubsection{AFib Detection}
Performance was measured by the Area under the Receiver Operating Characteristic (ROC-AUC), the Area under the Precision-Recall Curve (PR-AUC), and the F1 score. The PR-AUC is considered a better measure for imbalanced data.

\subsection{Implementation and Results}
\subsubsection{ECG Translation from PPG}

\begin{table}[h!]
\centering
\footnotesize
\caption{ECG translation performance of different models. The top three rows show models' performance on healthy subjects, while the fourth row shows the performance on both healthy and AFib subjects. If not specified, healthy subjects and clean signals is the default setting. The LSTM model \cite{tang2022robust} is subject-dependent, while the P2E-WGAN \cite{vo2021p2e} and our model are subject-independent.}
\label{tab:perf-tab}
\begin{tabular}{lccc}
% \toprule
 & Correlation & RMSE (mV) & SNR (dB) \\
\midrule
ADSSM &
  0.858 $\pm$ 0.174 &
  0.07 $\pm$ 0.047 &
  15.365 $\pm$ 11.053 \\
\midrule
ADSSM \\ w/o attention &
  0.823 $\pm$ 0.194 &
  0.08 $\pm$ 0.047 &
  13.013 $\pm$ 10.537 \\
\midrule
ADSSM \\ (healthy sub., \\ noisy sig.) &
  0.847 $\pm$ 0.174 &
  0.076 $\pm$ 0.049 &
  13.887 $\pm$ 10.58 \\
\midrule
ADSSM \\ (healthy \& \\AFib sub.) &
  0.804 $\pm$ 0.22 &
  0.078 $\pm$ 0.05 &
  12.261 $\pm$ 11.328 \\
\midrule
P2E-WGAN &
  0.773 $\pm$ 0.242 &
  0.091 $\pm$ 0.052 &
  9.616 $\pm$ 9.252 \\
\midrule
LSTM \\ (sub. dependent) &
  0.766 $\pm$ 0.234 &
  0.093 $\pm$ 0.053 &
  8.189 $\pm$ 9.560 \\
\bottomrule
\end{tabular}

\end{table}

Table \ref{tab:perf-tab} shows the performance of our model and compares it with other models in terms of means and standard deviations of $\rho$, RMSE and SNR. 
The correlation between the signals generated by our model and the reference signals is statistically strong, with a value $\rho$ of 0.858. Also, low values of RMSE (0.07) and high SNR (15.365) show strong similarities between them and reference ECG signals. When the attention mechanism is not applied on the input PPG, there is a notable decline in performance, with $\rho$ falling to 0.823, RMSE increasing to 0.08, and SNR decreasing to 13.013. This underscores the importance of the mechanism in providing relevant contexts for translation. The third row shows our model's performance on the noisy dataset. The negligible drop in metrics from 0.858 to 0.847 ($\rho$), 0.07 to 0.76 (RMSE), and 15.365 to 13.887 (SNR) demonstrates the robustness of our model. We attribute this to the probabilistic nature of the model, which better handles the measurement noise. As expected, the model performed worse on subjects with AFib due to the erratic patterns of the AFib signals (no visible P waves and an irregularly irregular QRS complex). In the next section, we show that the synthetic AFib signals are beneficial to the downstream detection task.

The P2E-WGAN model \cite{vo2021p2e}, a 1D deep convolutional generative adversarial network (4,064,769 parameters) for signal-to-signal translation, was recently proposed to translate PPG into ECG signals from a large number of subjects. P2E-WGAN achieved significantly lower performance than our model (645,466), requiring almost six times the parameters. Our model is less affected when data is scarce, which is common in healthcare. On the other hand, the LSTM model \cite{tang2022robust} is a deep recurrent neural network that was also recently proposed and built separately for each subject. The performance of our model, trained in a cross-subject setting, surpassed that of the LSTM model trained separately for each subject. These results prove the effectiveness and efficiency of our proposed sequential data structure. Further work with a larger number of subjects having AFib is needed to demonstrate that we can extend the model to new individuals. In addition, exploring strategies to manage the class imbalance problem \cite{johnson2019survey}, which arises from the fewer AFib records compared to the healthy ones, would be beneficial.

\begin{table}[ht!]
\centering
\footnotesize
\caption{AFib detection performance. The performance on the translated ECG is evaluated when the MINA model \cite{hong2019mina} is trained on real ECG but tested on synthetic ECG. The fusion performance is when the MINA model is extended to receive both real ECG and synthetic ECG inputs. x\% random time samples are omitted, simulating intermittent ECG recording, while synthetic ECG is always available.}
\label{tab:afib-tab}
\begin{tabular}{cccc}
& Real ECG & Translated ECG \\
\midrule
ROC-AUC & 0.995 $\pm$ 0.006 & 0.99 $\pm$ 0.004\\
PR-AUC & 0.987 $\pm$ 0.013 & 0.986 $\pm$ 0.007\\
F1 & 0.985 $\pm$ 0.009 & 0.944 $\pm$ 0.014\\
\midrule
Fusion & 30\% missing & 50\% missing & 70\% missing \\
\midrule
ROC-AUC & 0.992 $\pm$ 0.006 & 0.99 $\pm$ 0.006 & 0.99 $\pm$ 0.009 \\
PR-AUC & 0.986 $\pm$ 0.011 & 0.982 $\pm$ 0.012 & 0.981 $\pm$ 0.016 \\
F1 & 0.971 $\pm$ 0.01 & 0.969 $\pm$ 0.012 & 0.956 $\pm$ 0.046 \\ 
\bottomrule
\end{tabular}
\end{table}

\begin{figure*}[htb!] % add tb to control float placement
  \centering
  \begin{subfigure}[b]{0.45\textwidth}
    \includegraphics[width=\textwidth]{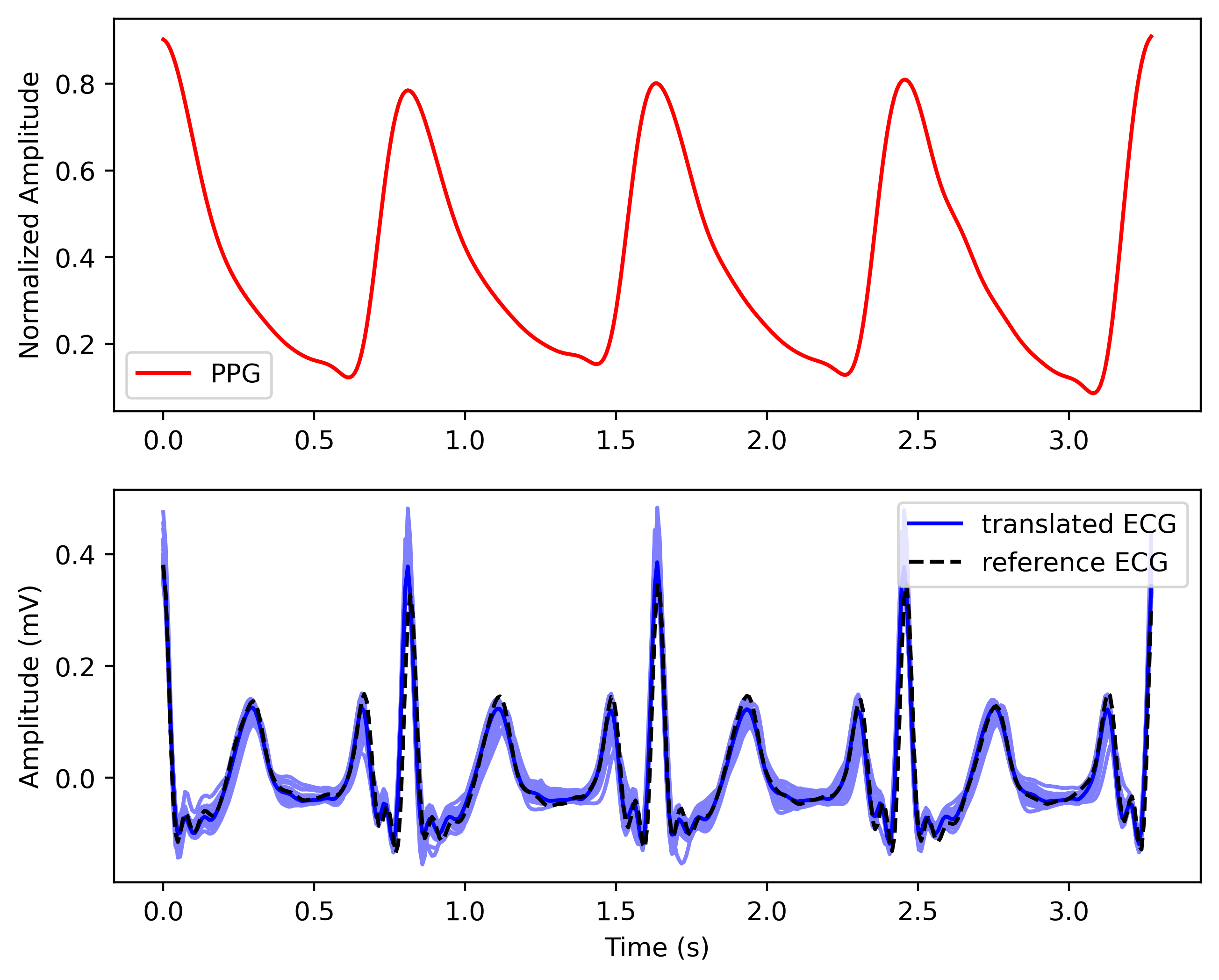}
    \caption{Clean input PPG}
    \label{fig:clean1}
  \end{subfigure}\hfil
  \begin{subfigure}[b]{0.45\textwidth}
    \includegraphics[width=\textwidth]{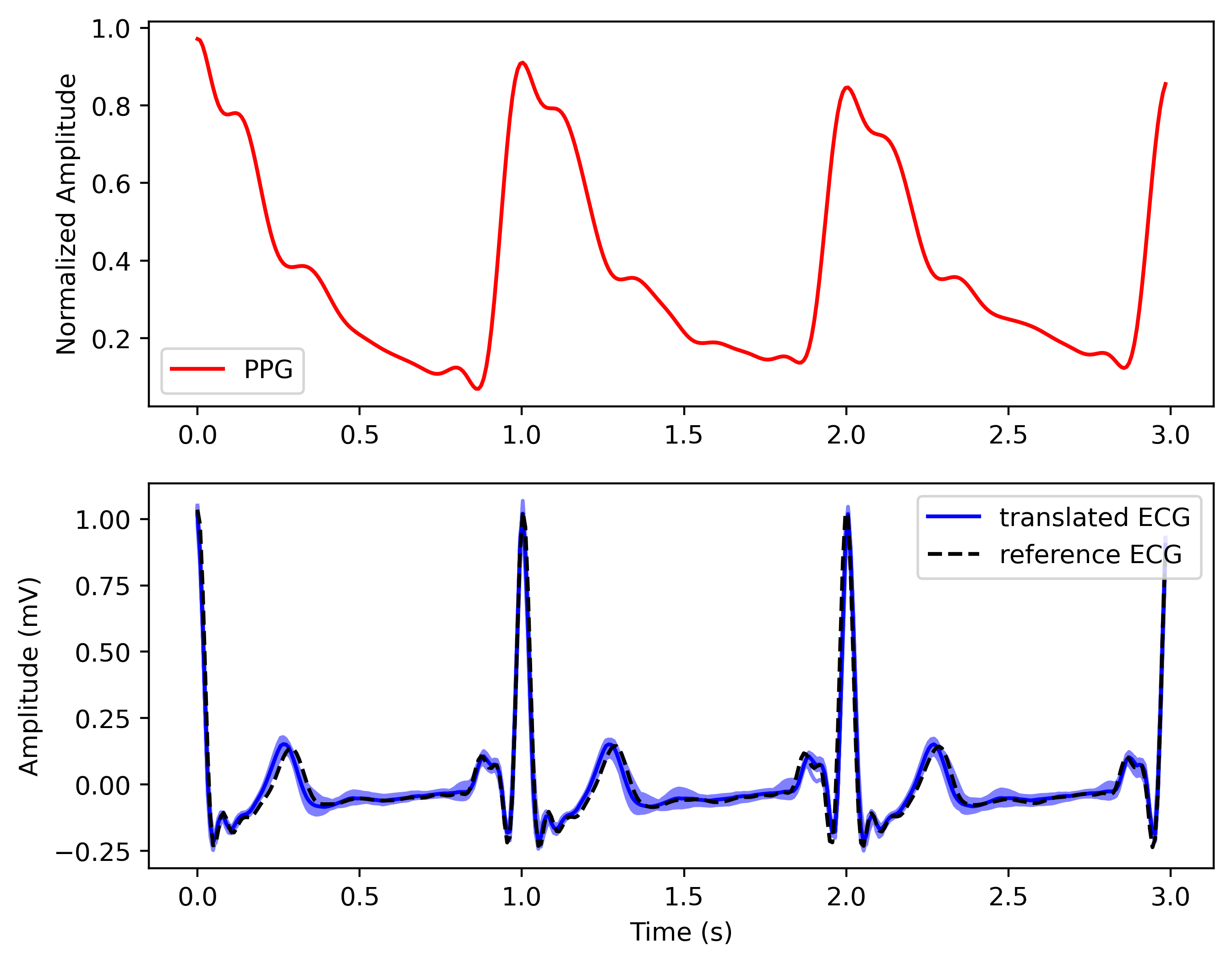}
    \caption{Clean input PPG}
    \label{fig:clean2}
  \end{subfigure}\par\medskip
  \begin{subfigure}[b]{0.45\textwidth}
    \includegraphics[width=\textwidth]{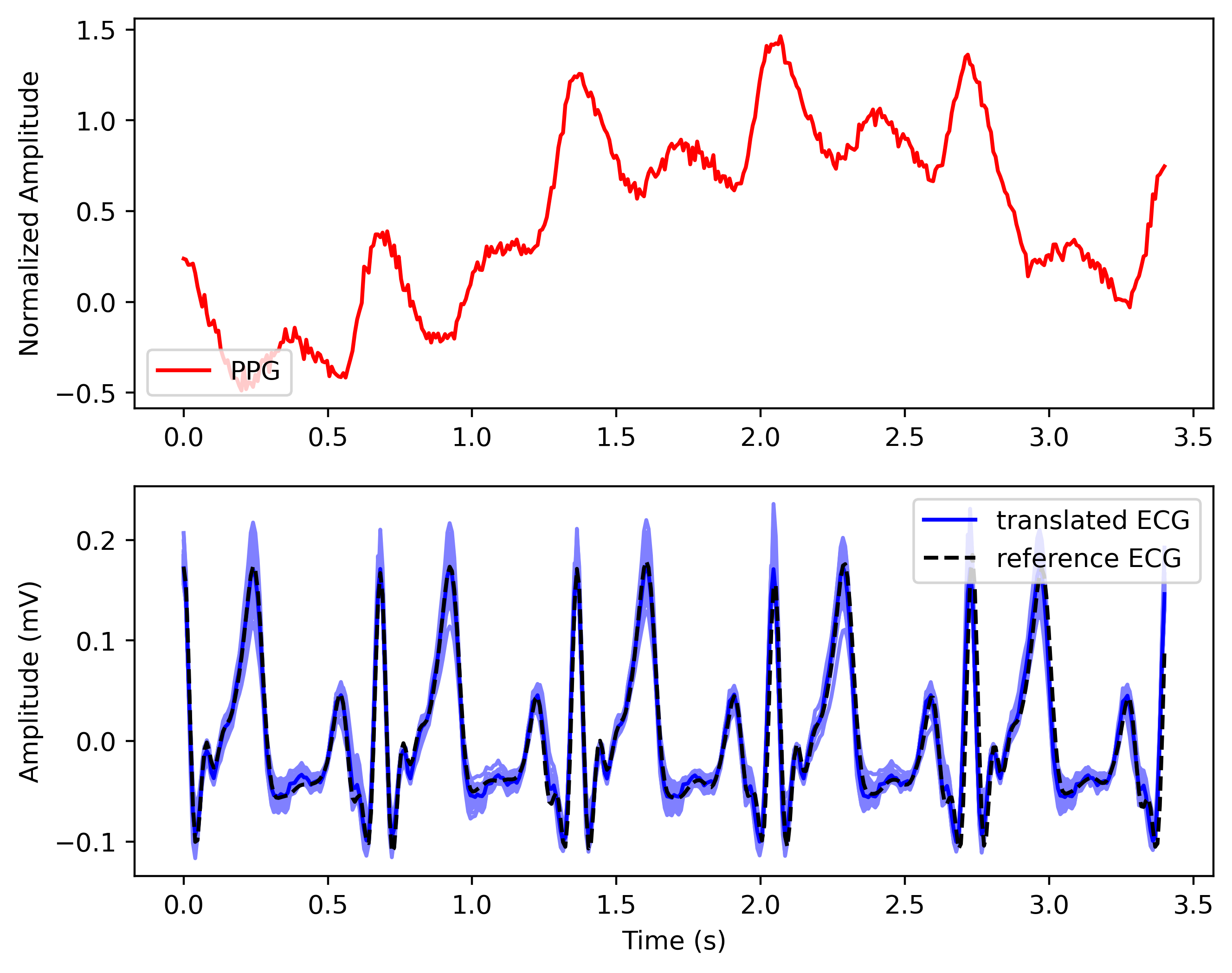}
    \caption{Noisy input PPG}
    \label{fig:noisy1}
  \end{subfigure}\hfil
  \begin{subfigure}[b]{0.45\textwidth}
    \includegraphics[width=\textwidth]{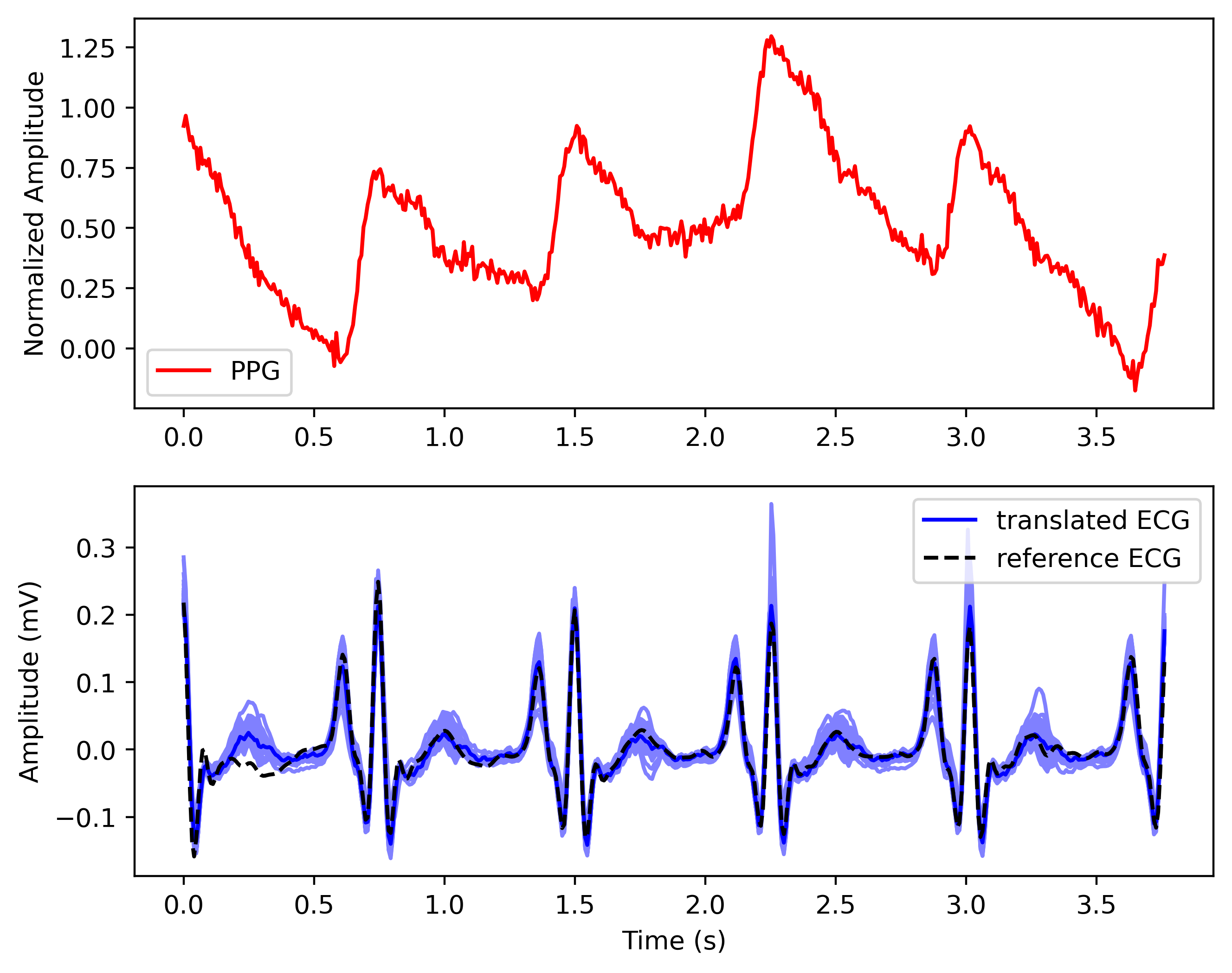}
    \caption{Noisy input PPG}
    \label{fig:noisy2}
    \end{subfigure}\par\medskip
  \begin{subfigure}[b]{0.45\textwidth}
    \includegraphics[width=\textwidth]{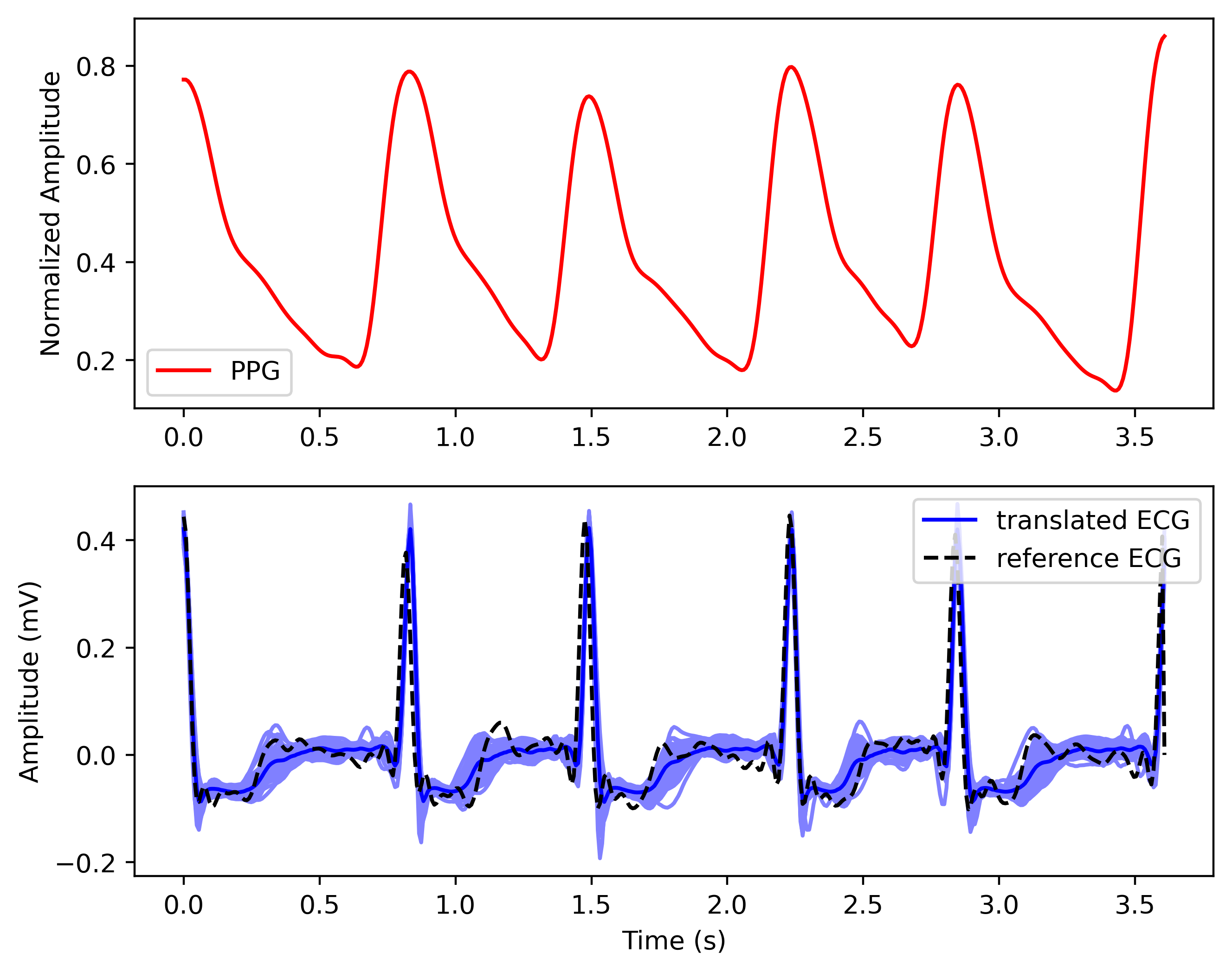}
    \caption{AFib input PPG}
    \label{fig:afib1}
  \end{subfigure}\hfil
  \begin{subfigure}[b]{0.45\textwidth}
    \includegraphics[width=\textwidth]{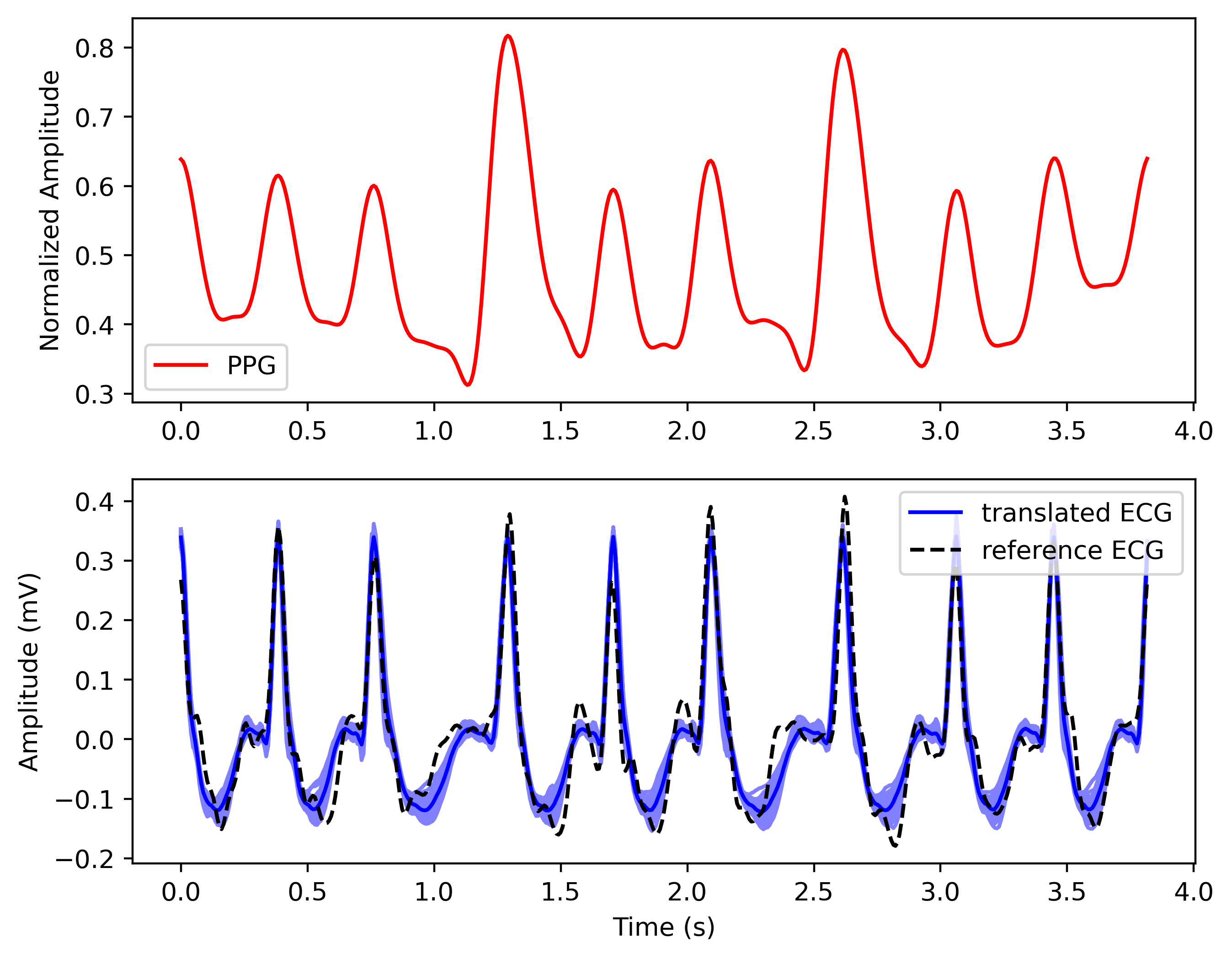}
    \caption{AFib input PPG}
    \label{fig:afib2}
  \end{subfigure}
  \caption{Examples of the translated ECG signals. In each subfigure: the top panel shows the input PPG waveform and the bottom panel shows the reconstructed ECG waveform compared with the reference waveform. The average ECG waveform (dark blue) of all possible pulses overlaid on each individual pulse (light blue).}
  \label{fig:examples}
\end{figure*}

In Figure \ref{fig:examples}, translated ECG waveforms are plotted with respect to the reference ECG waveforms of different heart rates. We can see that the model closely reconstructed the waveforms and maintained their essential properties, such as the missing P waves of the AFib ECG. In addition, we can be informed of the translation uncertainty by using a posterior on the latent embedding to propagate uncertainty from the embedding to the data. More specifically, with a distribution $p(\mathbf{z})$ on the latent feature our predictions will be $p_\theta\left(\mathbf{y} \mid \mathbf{x}\right)=\int p_{\theta_y}\left(\mathbf{y} \mid \mathbf{z}\right) p_{\theta_z}\left(\mathbf{z} \mid \mathbf{x}\right) d\mathbf{z}$. This would make the model more trustworthy and give patients and clinicians greater confidence in using it for medical diagnosis \cite{begoli2019need}. Future studies are expected to investigate methods to develop a fully Bayesian model and introduce a more flexible latent space \cite{tran2023fully, bendekgey2024unbiased}. Such advancements are advantageous in the medical field, particularly when data availability is limited or when uncertainty quantification and learning interpretable representations are essential.

\subsubsection{AFib Detection}

We evaluated the performance of our model on the benefits of the translated ECG for the AFib detection task. To do so, we used a state-of-the-art AFib detection model, Multilevel Knowledge-Guided Attention (MINA) \cite{hong2019mina}, trained on real ECG signals, each of 10 s, and tested against synthetic. It should be noted that any pre-trained AFib detection model can be used in our pipeline. Table \ref{tab:afib-tab} reveals the mean detection performance of the model in the translated ECG that is close to that of the real ECG, ROC-AUC of 0.99 vs. 0.995, PR-AUC of 0.986 vs. 0.987, and F1 of 0.944 vs. 0.985. This implies that our model allows for the combined advantages of ECG's rich knowledge base and PPG's continuous measurement.

Furthermore, we extended the ability of the MINA model to receive real and translated ECG signals by incorporating the translated frequency channels into the model. In this scenario, both ECG and PPG signals can be measured simultaneously. This setting requires retraining of the MINA model on the fused real and synthetic ECG signal data set. To simulate the real-life setting where ECG measurement is intermittent while PPG input is continuous, we randomly zeroed out time samples with different probabilities: 30\%, 50\%, and 70\%. As shown in the bottom results of Table \ref{tab:afib-tab}, the performance remains almost unchanged in the fusion mode across the omission thresholds. Additionally, the model learns to utilize the sparse real ECG to marginally improve performance against only the translated ECG.

\section{Conclusion}

In this work, we present a novel attention-based deep state-space model to generate ECG waveforms with PPG signals as input. The results demonstrate that our model has the potential to provide a paradigm shift in telemedicine by bringing about ECG-based clinical diagnoses of heart disease via simple PPG assessment through wearable devices. 
Our model, trained on a small and noisy dataset, achieves an average Pearson's correlation of 0.847, RMSE of 0.076 mV, and SNR of 13.887 dB, demonstrating the efficacy of our approach. Significantly, our model enables the AFib monitoring capability in a continuous setting, assisting a state-of-the-art AFib detection model to achieve a PR-AUC of 0.986. Being a lightweight method also facilitates its deployment on resource-constrained devices.
% In our future work, we plan to further validate the proposed method with other ECG and PPG datasets that contain noisy PPG signals where the source of noise is from daily activities.
In our future work, we aim to validate the generalizability of the model with other pairs of physiological signals. Our method allows for the screening and early detection of cardiovascular diseases in the home environment, saving money and labor, while supporting society in unusual pandemic situations.
% \FloatBarrier

\bibliographystyle{IEEEbib}
\bibliography{refs}

\begin{thebibliography}{10}

\bibitem{ref:Allen}
J.~Allen,
\newblock ``Photoplethysmography and its application in clinical physiological measurement,''
\newblock {\em Physiol Meas.}, vol. 28, no. 3, pp. 1--39, 2007.

\bibitem{ref:sudden}
sudden,
\newblock ``Sudden death in young people: Heart problems often blamed,'' Retrieved on September 2020,
\newblock \url{https://www.mayoclinic.org/diseases-conditions/sudden-cardiac-arrest/in-depth/sudden-death/art-20047571}.

\bibitem{rosiek2016risk}
Anna Rosiek and Krzysztof Leksowski,
\newblock ``The risk factors and prevention of cardiovascular disease: the importance of electrocardiogram in the diagnosis and treatment of acute coronary syndrome,''
\newblock {\em Therapeutics and clinical risk management}, vol. 12, pp. 1223, 2016.

\bibitem{olier2021machine}
Ivan Olier, Sandra Ortega-Martorell, Mark Pieroni, and Gregory~YH Lip,
\newblock ``How machine learning is impacting research in atrial fibrillation: implications for risk prediction and future management,''
\newblock {\em Cardiovascular Research}, vol. 117, no. 7, pp. 1700--1717, 2021.

\bibitem{reisner2008utility}
Andrew Reisner, Phillip~A Shaltis, Devin McCombie, and H~Harry Asada,
\newblock ``Utility of the photoplethysmogram in circulatory monitoring,''
\newblock {\em Anesthesiology: The Journal of the American Society of Anesthesiologists}, vol. 108, no. 5, pp. 950--958, 2008.

\bibitem{banerjee2014photoecg}
Rohan Banerjee, Aniruddha Sinha, Anirban~Dutta Choudhury, and Aishwarya Visvanathan,
\newblock ``Photoecg: Photoplethysmographyto estimate ecg parameters,''
\newblock in {\em 2014 IEEE International Conference on Acoustics, Speech and Signal Processing (ICASSP)}. IEEE, 2014, pp. 4404--4408.

\bibitem{zhu2019ecg}
Qiang Zhu, Xin Tian, Chau-Wai Wong, and Min Wu,
\newblock ``Ecg reconstruction via ppg: A pilot study,''
\newblock in {\em 2019 IEEE EMBS International Conference on Biomedical \& Health Informatics (BHI)}. IEEE, 2019, pp. 1--4.

\bibitem{tian2020cross}
Xin Tian, Qiang Zhu, Yuenan Li, and Min Wu,
\newblock ``Cross-domain joint dictionary learning for ecg reconstruction from ppg,''
\newblock in {\em ICASSP 2020-2020 IEEE International Conference on Acoustics, Speech and Signal Processing (ICASSP)}. IEEE, 2020, pp. 936--940.

\bibitem{vo2021p2e}
Khuong Vo, Emad~Kasaeyan Naeini, Amir Naderi, Daniel Jilani, Amir~M Rahmani, Nikil Dutt, and Hung Cao,
\newblock ``P2e-wgan: Ecg waveform synthesis from ppg with conditional wasserstein generative adversarial networks,''
\newblock in {\em Proceedings of the 36th Annual ACM Symposium on Applied Computing}, 2021, pp. 1030--1036.

\bibitem{sarkar2021cardiogan}
Pritam Sarkar and Ali Etemad,
\newblock ``Cardiogan: Attentive generative adversarial network with dual discriminators for synthesis of ecg from ppg,''
\newblock in {\em Proceedings of the AAAI Conference on Artificial Intelligence}, 2021, vol.~35, pp. 488--496.

\bibitem{chiu2020reconstructing}
Hong-Yu Chiu, Hong-Han Shuai, and Paul C-P Chao,
\newblock ``Reconstructing qrs complex from ppg by transformed attentional neural networks,''
\newblock {\em IEEE Sensors Journal}, vol. 20, no. 20, pp. 12374--12383, 2020.

\bibitem{lee2020stint}
Tao-Yi Lee, Khuong Vo, Wongi Baek, Michelle Khine, and Nikil Dutt,
\newblock ``Stint: selective transmission for low-energy physiological monitoring,''
\newblock in {\em Proceedings of the ACM/IEEE International Symposium on Low Power Electronics and Design}, 2020, pp. 115--120.

\bibitem{hong2020opportunities}
Shenda Hong, Yuxi Zhou, Junyuan Shang, Cao Xiao, and Jimeng Sun,
\newblock ``Opportunities and challenges of deep learning methods for electrocardiogram data: A systematic review,''
\newblock {\em Computers in biology and medicine}, vol. 122, pp. 103801, 2020.

\bibitem{krishnan2015deep}
Rahul~G Krishnan, Uri Shalit, and David Sontag,
\newblock ``Deep kalman filters,''
\newblock {\em arXiv preprint arXiv:1511.05121}, 2015.

\bibitem{krishnan2017structured}
Rahul Krishnan, Uri Shalit, and David Sontag,
\newblock ``Structured inference networks for nonlinear state space models,''
\newblock in {\em Proceedings of the AAAI Conference on Artificial Intelligence}, 2017, vol.~31.

\bibitem{bahdanau2014neural}
Dzmitry Bahdanau, Kyunghyun Cho, and Yoshua Bengio,
\newblock ``Neural machine translation by jointly learning to align and translate,''
\newblock {\em arXiv preprint arXiv:1409.0473}, 2014.

\bibitem{kingma2013auto}
Diederik~P Kingma and Max Welling,
\newblock ``Auto-encoding variational bayes,''
\newblock {\em arXiv preprint arXiv:1312.6114}, 2013.

\bibitem{sohn2015learning}
Kihyuk Sohn, Honglak Lee, and Xinchen Yan,
\newblock ``Learning structured output representation using deep conditional generative models,''
\newblock {\em Advances in neural information processing systems}, vol. 28, 2015.

\bibitem{chung2014empirical}
Junyoung Chung, Caglar Gulcehre, KyungHyun Cho, and Yoshua Bengio,
\newblock ``Empirical evaluation of gated recurrent neural networks on sequence modeling,''
\newblock {\em arXiv preprint arXiv:1412.3555}, 2014.

\bibitem{kingma2014adam}
Diederik~P Kingma and Jimmy Ba,
\newblock ``Adam: A method for stochastic optimization,''
\newblock {\em arXiv preprint arXiv:1412.6980}, 2014.

\bibitem{moody2020mimic}
B~Moody, G~Moody, M~Villarroel, G~Clifford, and I~Silva,
\newblock ``Mimic-iii waveform database matched subset,''
\newblock {\em MIMIC-III Waveform Database Matched Subset v1. 0}, 2020.

\bibitem{johnson2016mimic}
Alistair~EW Johnson, Tom~J Pollard, Lu~Shen, Li-wei~H Lehman, Mengling Feng, Mohammad Ghassemi, Benjamin Moody, Peter Szolovits, Leo Anthony~Celi, and Roger~G Mark,
\newblock ``Mimic-iii, a freely accessible critical care database,''
\newblock {\em Scientific data}, vol. 3, no. 1, pp. 1--9, 2016.

\bibitem{tang2022robust}
Qunfeng Tang, Zhencheng Chen, Yanke Guo, Yongbo Liang, Rabab Ward, Carlo Menon, and Mohamed Elgendi,
\newblock ``Robust reconstruction of electrocardiogram using photoplethysmography: A subject-based model,''
\newblock {\em Frontiers in Physiology}, p. 645, 2022.

\bibitem{van2019heartpy}
Paul Van~Gent, Haneen Farah, Nicole Van~Nes, and Bart Van~Arem,
\newblock ``Heartpy: A novel heart rate algorithm for the analysis of noisy signals,''
\newblock {\em Transportation research part F: traffic psychology and behaviour}, vol. 66, pp. 368--378, 2019.

\bibitem{van2019analysing}
Paul van Gent, Haneen Farah, Nicole van Nes, and Bart van Arem,
\newblock ``Analysing noisy driver physiology real-time using off-the-shelf sensors: Heart rate analysis software from the taking the fast lane project,''
\newblock {\em Journal of Open Research Software}, vol. 7, no. 1, 2019.

\bibitem{lu2009comparison}
Guohua Lu, F~Yang, JA~Taylor, and John~F Stein,
\newblock ``A comparison of photoplethysmography and ecg recording to analyse heart rate variability in healthy subjects,''
\newblock {\em Journal of medical engineering \& technology}, vol. 33, no. 8, pp. 634--641, 2009.

\bibitem{aschbacher2020atrial}
Kirstin Aschbacher, Defne Yilmaz, Yaniv Kerem, Stuart Crawford, David Benaron, Jiaqi Liu, Meghan Eaton, Geoffrey~H Tison, Jeffrey~E Olgin, Yihan Li, et~al.,
\newblock ``Atrial fibrillation detection from raw photoplethysmography waveforms: A deep learning application,''
\newblock {\em Heart rhythm O2}, vol. 1, no. 1, pp. 3--9, 2020.

\bibitem{johnson2019survey}
Justin~M Johnson and Taghi~M Khoshgoftaar,
\newblock ``Survey on deep learning with class imbalance,''
\newblock {\em Journal of Big Data}, vol. 6, no. 1, pp. 1--54, 2019.

\bibitem{hong2019mina}
Shenda Hong, Cao Xiao, Tengfei Ma, Hongyan Li, and Jimeng Sun,
\newblock ``Mina: multilevel knowledge-guided attention for modeling electrocardiography signals,''
\newblock {\em arXiv preprint arXiv:1905.11333}, 2019.

\bibitem{begoli2019need}
Edmon Begoli, Tanmoy Bhattacharya, and Dimitri Kusnezov,
\newblock ``The need for uncertainty quantification in machine-assisted medical decision making,''
\newblock {\em Nature Machine Intelligence}, vol. 1, no. 1, pp. 20--23, 2019.

\bibitem{tran2023fully}
Ba-Hien Tran, Babak Shahbaba, Stephan Mandt, and Maurizio Filippone,
\newblock ``Fully bayesian autoencoders with latent sparse gaussian processes,''
\newblock in {\em International Conference on Machine Learning}. PMLR, 2023, pp. 34409--34430.

\bibitem{bendekgey2024unbiased}
Henry~C Bendekgey, Gabe Hope, and Erik Sudderth,
\newblock ``Unbiased learning of deep generative models with structured discrete representations,''
\newblock {\em Advances in Neural Information Processing Systems}, vol. 36, 2024.

\end{thebibliography}

\end{document}